\documentclass[letterpaper]{article} 
\usepackage{aaai2026}  
\usepackage{times}  
\usepackage{helvet}  
\usepackage{courier}  
\usepackage[hyphens]{url}  
\usepackage{graphicx} 
\usepackage{natbib}  
\usepackage{caption} 
\usepackage{algorithm}
\usepackage{algorithmic}

\usepackage{newfloat}
\usepackage{listings}
\DeclareCaptionStyle{ruled}{labelfont=normalfont,labelsep=colon,strut=off} 
\floatstyle{ruled}
\newfloat{listing}{tb}{lst}{}
\floatname{listing}{Listing}
\usepackage{amssymb}
\usepackage{amsmath} 
\usepackage{booktabs} 

\usepackage{pifont}
\newcommand{\cmark}{\ding{51}}   
\newcommand{\xmark}{\ding{55}}   

\usepackage{nameref}
\usepackage{array,booktabs,multirow,siunitx}
\newcolumntype{C}[1]{>{\centering\arraybackslash}m{#1}}
\newcolumntype{L}[1]{>{\raggedright\arraybackslash}p{#1}}
\usepackage{enumitem}
\usepackage{rotating}
\usepackage{adjustbox}

\usepackage{multirow}
\usepackage{makecell}
\usepackage{subcaption}

\title{CLIP-FTI: Fine-Grained Face Template Inversion via\\CLIP-Driven Attribute Conditioning}
\author{
    Longchen Dai,
    Zixuan Shen$^*$,
    Zhiheng Zhou,
    Peipeng Yu,
    Zhihua Xia\footnote{Zhihua Xia and Zixuan Shen are co-corresponding authors.}
}
\affiliations{
    College of Cyber Security, Jinan University\\
    \{LongchenDai, jnuzzh\}@stu2023.jnu.edu.cn, \{m15195901120, ypp865, xia\_zhihua\}@163.com

}

\usepackage{bibentry}

\begin{document}

\maketitle

\begin{abstract}
Face recognition systems store face templates for efficient matching. Once leaked, these templates pose a threat: inverting them can yield photorealistic surrogates that compromise privacy and enable impersonation. Although existing research has achieved relatively realistic face template inversion, the reconstructed facial images exhibit over-smoothed facial-part attributes (eyes, nose, mouth) and limited transferability. To address this problem, we present CLIP-FTI, a CLIP-driven fine-grained attribute conditioning framework for face template inversion. Our core idea is to use the CLIP model to obtain the semantic embeddings of facial features, in order to realize the reconstruction of specific facial feature attributes. Specifically, facial feature attribute embeddings extracted from CLIP are fused with the leaked template via a cross-modal feature interaction network and projected into the intermediate latent space of a pretrained StyleGAN. The StyleGAN generator then synthesizes face images with the same identity as the templates but with more fine-grained facial feature attributes. Experiments across multiple face recognition backbones and datasets show that our reconstructions \emph{(i) achieve higher identification accuracy and attribute similarity, (ii) recover sharper component-level attribute semantics, and (iii) improve cross-model attack transferability compared to prior reconstruction attacks}. To the best of our knowledge, ours is the first method to use additional information besides the face template attack to realize face template inversion and obtains SOTA results.
\end{abstract}


\section{Introduction}

Face recognition (FR) systems ubiquitously store facial templates (i.e., compact deep embeddings extracted during enrollment) for efficient verification and identification across applications such as smartphone unlocking and border control. Once leaked, these templates can be used to attack FR systems: face template inversion (FTI) attacks can reconstruct photorealistic facial surrogates that simultaneously (i) compromise user privacy (exposing soft biometric attributes such as age, gender, ethnicity, or skin tone) and (ii) enable downstream impersonation attacks. Unlike most attacks on FR systems that threaten the security of the system \cite{biggio2015adversarial,deb2020advfaces} and biometric privacy \cite{galbally2014biometric,galbally2010vulnerability,hadid2015biometrics, marcel2023handbook,marcel2019handbook}. FTI directly undermines both privacy and security by recovering an identity's image from its leaked template and thus requires further study.

Early FTI attacks attempt employed custom encoder-decoder networks \cite{mai2018reconstruction} that directly mapped templates back to pixel space, but were constrained by limited generative capacity and low visual fidelity. Recent methods leverage powerful pretrained generative adversarial networks (GANs) \cite{otroshi2023face}, first projecting facial templates into an intermediate latent space, and then synthesizing higher-resolution face images. While these GAN-assisted approaches markedly improve realism, they predominantly rely on a single leaked face template. As a result, reconstructions often exhibit over-smoothed facial feature attributes (periocular shape, nasal ridge, lip contours).

Text-conditioned image synthesis provides a new approach for the fine-grained recovery of facial template attributes. Relevant studies shows that semantic descriptors can steer generative models toward attribute-faithful refinements without eroding global structure. Inspired by this, we hypothesize that augmenting leaked templates with semantic attribute embeddings can (i) enrich reconstruction details and (ii) elevate cross-model transferability.

We introduce \textbf{CLIP-FTI} (Face Template Inversion via Fine-Grained Attribute Conditioning), a CLIP-driven face template inversion framework. Given a leaked facial template \(t\), we derive facial semantic attribute embeddings using pretrained CLIP model by matching the template's paired image against a curated bank of textual prompts that describe fine facial feature attributes (e.g., eye-shape variants, nasal bridge traits, lip fullness descriptors). These attribute embeddings are aggregated into a single representation \(s\).
Then we use two learned mapping sub-modules of CLIP-FTI:  
(1) a \emph{template→attribute alignment} (TAA) adapter that predicts the CLIP-based attribute embedding \(s\) from the leaked template \( t \); and (2) a \emph{fusion→latent projector} that first fuses \(t\) and \(s\) into a joint code \(f\) (via cross-modal attention) and then maps \(f\) to a StyleGAN latent code \(\mathcal{W}\). The frozen StyleGAN generator subsequently synthesizes an identity-consistent, attribute-faithful face image. By conditioning on CLIP-derived facial attribute semantic embedding in a single forward pass, CLIP-FTI compensates for detail defects of prior template-only inversions, yielding higher identification accuracy, sharper attribute similarity, and stronger cross-model transferability.

To elaborate on the contributions of this paper, we list them hereunder:

\begin{itemize}
    \item To our best knowledge, we make the first attempt to explore face template inversion using CLIP-driven semantic attribute embeddings, mitigating over-smoothed facial feature attribute details prevalent in prior method. Specifically, we design a cross-modal feature interaction network that fuses leaked templates with CLIP attribute embeddings and maps the fused representation into StyleGAN's intermediate latent space, explicitly addressing the identity-detail trade-off.
    \item We introduce a supervised mapping from facial templates to the corresponding attribute embeddings, enhancing semantic compatibility and improving downstream reconstruction fidelity.
    \item Experiments across multiple FR backbones and datasets show consistent gains in attribute similarity, identification accuracy, and cross-model attack transferability over state-of-the-art inversion attacks.
\end{itemize}

\section{Related Work \& Preliminary}

\label{sec:related}
\subsection{Face Template Inversion}  
The early facial template inversion attacks typically trained bespoke encoder-decoder networks that map a leaked template to a low-resolution image without any generative prior. Mai \emph{et al.}~\cite{mai2018reconstruction} and Cole \emph{et al.}~\cite{cole2017synthesizing} jointly minimised pixel and perceptual loss, but the absence of a strong generator led to blurry faces and poor high-frequency detail, even in white-box settings. With the advent of StyleGAN, most subsequent works learned a \emph{global} mapping from a template to the generator latent space. Dong \emph{et al.}~\cite{dong2021towards} and Otroshi \emph{et al.}~\cite{otroshi2023face} regress the template to StyleGAN2/3 latent space via an MLP, whereas Shahreza \emph{et al.}~\cite{shahreza2024template, shahreza2023comprehensive} exploit synthetic pairs or a NeRF-based EG3D backbone. A complementary line searches the latent space per target—e.g. simulated annealing~\cite{vendrow2021realistic} or genetic algorithms~\cite{dong2023reconstruct}. These black-box attacks can recover identity but require hundreds of forward passes per image, making them computationally prohibitive. Although these methods inherit the visual quality of modern generators, relying on a single latent vector produces over-smoothed facial feature attributes (eyes, nose, and mouth), limiting both realism and attack transferability.

\subsection{Vision--Language Models and CLIP} 
Contrastive Language-Image Pre-training (CLIP)~\cite{radford2021learning} aligns images and text in a shared embedding space by learning from millions of (image, caption) pairs. Its strong zero-shot transfer has spurred a series of generative applications, including prompt-based image editing (StyleCLIP~\cite{patashnik2021styleclip}), domain adaptation (StyleGAN-NADA~\cite{gal2022stylegan}), and text-guided 3D synthesis (DreamFusion~\cite{poole2022dreamfusion}). CLIP2StyleGAN~\cite{DBLP:journals/corr/abs-2112-05219} uses CLIP to discover and name disentangled semantic directions and then edits faces by moving StyleGAN latents along those directions; Arc2Face~\cite{paraperas2024arc2face} replaces text conditioning with an ArcFace ID embedding to synthesize high-quality, ID-consistent faces with a diffusion backbone.

In contrast, our goal is a targeted face template inversion (FTI) attack. Unlike general CLIP-guided editing or ID-consistent synthesis, we use CLIP to extract fine-grained facial-attribute embeddings (eyes, nose, mouth) and fuse them with the leaked template in a single pass. This guidance enforces text-semantic fidelity and yields high-resolution reconstructions that retain local details while fooling the target FR model. Prior template attacks often produce over-smoothed faces; our attribute-guided design mitigates this and improves attack efficacy. A concise comparison appears in Table \ref{tab:related-works}.

\begin{table}[t]
\centering
\setlength{\tabcolsep}{4pt}
\small
\begin{tabular}{lccccc}
\toprule
\textbf{Method} & \textbf{Year} & \textbf{Gen./Res.} & \textbf{FG} & \textbf{1-shot} \\
\midrule
Cole \textit{et al.} & 2017 & CNN / 128 & \xmark & \cmark \\
Mai \textit{et al.} & 2018 & CNN / 128 & \xmark & \cmark \\
FaceID-GAN & 2021 & SG2 / 512 & \xmark & \cmark \\
Otroshi \textit{et al.} & 2023 & SG3 / 1024 & \xmark & \cmark \\
Vendrow \textit{et al.} & 2021 & SG2 / 1024 & \cmark & \xmark \\
Dong \textit{et al.} & 2023 & SG2 / 1024 & \cmark & \xmark \\
\textbf{CLIP-FTI (ours)} & 2025 & SG3 / 1024 & \cmark & \cmark \\
\bottomrule
\end{tabular}
\caption{Comparison with prior facial template inversion attacks. "FG" = recovers fine-grained attribute details; "1-shot" = single forward pass at inference; "SG2/3" = StyleGAN2/3.}
\label{tab:related-works}
\vspace{-0.6em}
\end{table}

\subsection{Preliminary}

\noindent\textbf{StyleGAN.} StyleGAN~\cite{otroshi2023face,karras2019style,karras2020analyzing} employs a mapping network \(f: \mathcal{Z} \rightarrow \mathcal{W}\) that transforms an input noise vector \(z \sim P_Z\) (e.g., Gaussian) into an intermediate latent code \(w = f(z)\). This disentangled latent space \(\mathcal{W}\) is fed (via per-layer affine transformations) into the synthesis network to modulate convolutional weights (style modulation). Hierarchical semantics emerge: early layers control coarse pose and face shape, middle layers affect facial components (eyes, nose, mouth), and later layers refine texture and color micro-details. 

\noindent\textbf{CLIP.} CLIP~\cite{radford2021learning} jointly trains an image encoder \(E_I(\cdot)\) and a text encoder \(E_T(\cdot)\) on large-scale image-text pairs using a symmetric contrastive loss. Given a mini-batch \(\{(x_i, d_i)\}_{i=1}^B\), where \(x_i\) is an image and \(d_i\) is its paired textual description, the encoded features are \(\ell_2\)-normalized and used to form a temperature-scaled cosine-similarity matrix. Cross-entropy then encourages each \(x_i\) to match only its corresponding \(d_i\) (and vice versa) while pushing apart mismatches. This yields a shared embedding space in which semantically related images and texts lie nearby. Empirically, CLIP captures a broad range of visual concepts, including fine facial feature attributes (e.g., eye shape, brow thickness, nose-bridge height, lip fullness, skin-tone variations), despite not being explicitly trained for biometrics.

\section{Method}
In this section, we formalize the face template inversion (FTI) threat scenario, and introduce our CLIP-driven fine-grained attribute conditioning model \textbf{CLIP-FTI}.

\begin{figure*}[t]
  \centering
  \includegraphics[width=0.8\textwidth]{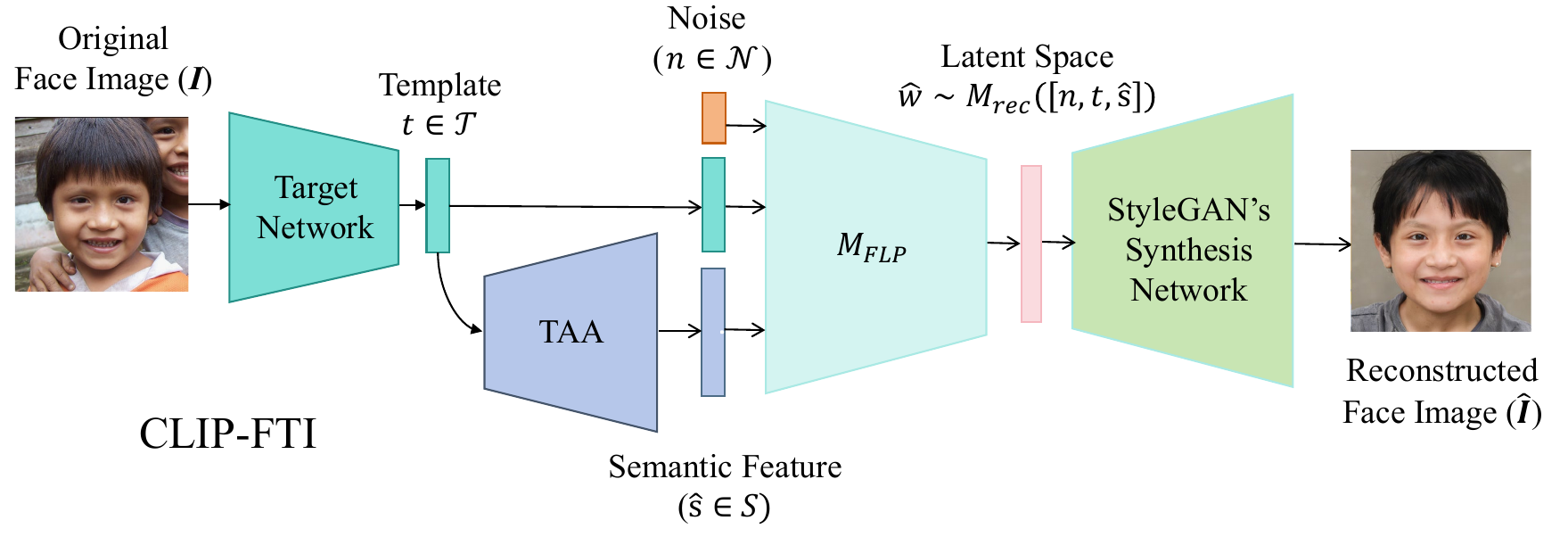}
  \caption{Overview of \textbf{CLIP-FTI}. An original image \(I\) yields a facial template \(t\) (ArcFace/ElasticFace). A template$\to$semantic alignment adapter \(TAA\) learns to predict \(\hat{s}\) from \(t\). A fusion mapping \(M_{\text{FLP}}\) takes noise \(n\), \(t\), and \(\hat{s}\) at attack time and outputs \(\hat{w} \in \mathcal{W}\); the frozen StyleGAN3 synthesis network synthesizes a reconstructed image \(\hat{I}=G(\hat{w})\).}
  \label{fig:framework}
\end{figure*}

\begin{figure}[t]
  \centering
  \includegraphics[width=1\linewidth]{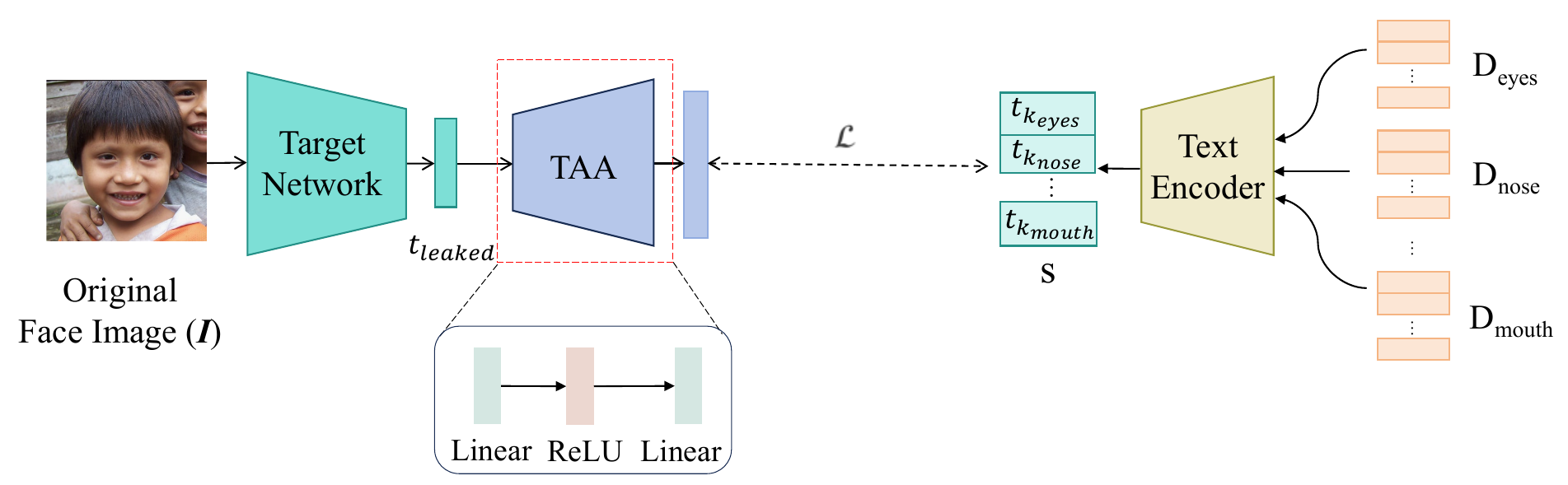}
  \caption{Training the template$\to$attribute alignment adapter \(TAA\). The adapter learns to predict the aggregated semantic embedding \(s\) from the facial template \(t\) using MSE and cosine loss.}
  \label{fig:TAA}
\end{figure}

\begin{figure*}[t]
  \centering
  \includegraphics[width=0.80\textwidth]{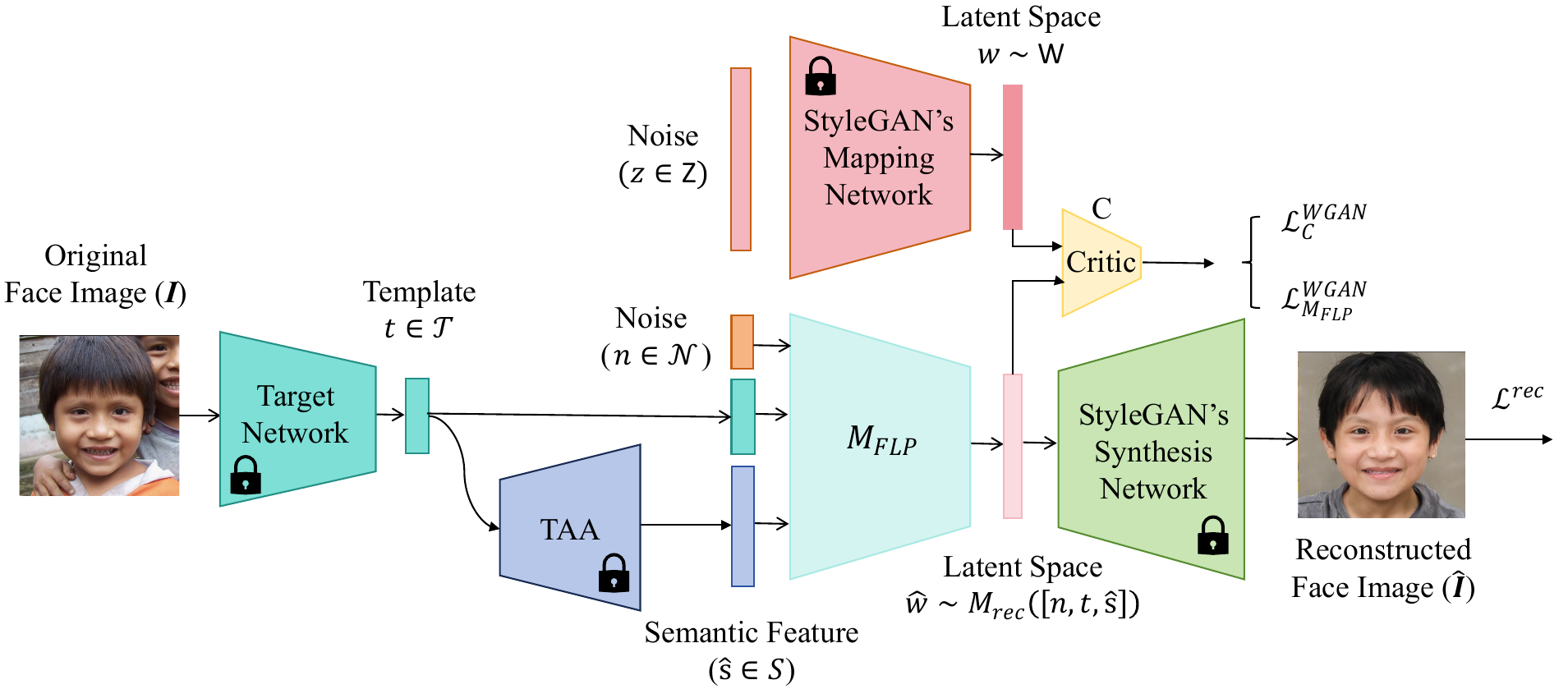}
  \caption{
    Training of the fusion-latent projector \(M_{FLP}\) and critic.
    A leaked template \(t\), its TAA-predicted attribute embedding \(\hat{s}\), and noise \(n\) are fused by \(M_{FLP}\) to produce a latent \(\hat{w}\); the critic aligns \(\hat{w}\) with the StyleGAN latent prior, while reconstruction losses between the generated image \(\hat{I}=G(\hat{w})\) and the original image \(I\) further supervise \(M_{FLP}\).
}

  \label{fig:fusion2latent}
\end{figure*}

\begin{figure}[t]
  \centering
  \includegraphics[width=1\linewidth]{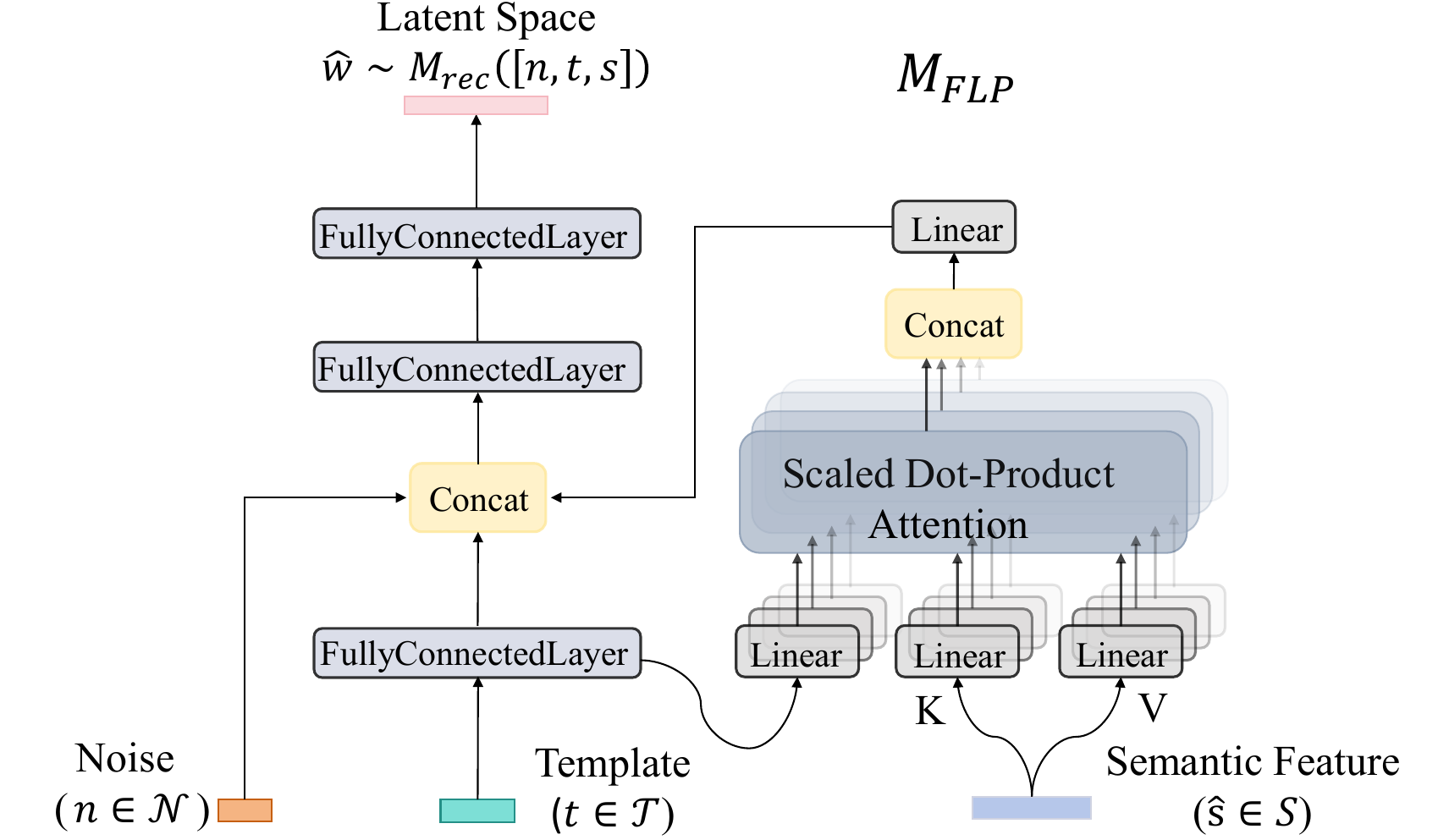}  
  \caption{Internal architecture of \(M_{\text{FLP}}\): three branches (noise \(n\), template \(t\), semantic embeddings \(\hat{s}\)) → attention fusion → MLP → latent \(w \in W\).}
  \label{fig:FLP}
\end{figure}

\noindent\textbf{Attack scenario.} We consider a FTI attack against a FR system based on the following scenario:
\begin{itemize}
    \item \textbf{Adversary's goal:} The adversary seeks to reconstruct a face image from a leaked facial template, and use the reconstructed face image to enter the target FR system.
    \item \textbf{Adversary's knowledge:} The adversary is assumed has access to a facial template, which was leaked from the database of the target's FR system. But lacks knowledge of the internal structure of the FR model (blackbox) and can only interact with the model through external interfaces (such as SDKs, APIs, etc.) for any face image.
    \item \textbf{Adversary's capability:} The adversary is able to utilize the reconstructed face image to access the target face recognition system. For simplicity, it is assumed that the adversary can inject the reconstructed face image as a query into the target system.
    \item \textbf{Adversary's approach:} The adversary can train a face reconstruction model to invert facial templates and reconstruct underlying face images. Then, the adversary can use reconstructed face images to inject as a query to the target FR system, to enter that system.
\end{itemize}

\noindent\textbf{CLIP-FTI.} To perform fine-grained face template inversion, we introduce \textbf{CLIP-FTI}, a CLIP-driven attribute conditioning framework for face template inversion. During \emph{training} stage, we assume access to face images and their corresponding recognition templates (extracted by a surrogate FR model) so that we can (i) extract CLIP-based semantic attribute embeddings and (ii) learn two mappings: a template$\to$attribute alignment (TAA) adapter and a fusion$\to$latent projector (\(M_{\text{FLP}}\)). During the \emph{attack} (inference) stage, only a leaked template \(t\) is available; the trained TAA adapter predicts a semantic attribute embedding \(\hat{s}\), which together with a noise vector \(z\) is fused to produce a latent code \( \hat{w} \in \mathcal{W} \); the frozen StyleGAN3 generator then synthesizes the reconstructed face \(\hat{I}=G(\hat{w})\). Figure~\ref{fig:framework} summarizes the pipeline.

\subsection{TAA Adapter}

\noindent\textbf{Facial feature attribute prompt matching.}
To enable more detailed semantic alignment, we divide the face into multiple regions—such as eyes, nose, and mouth—and predefine a set of textual descriptions for each region:
\[
\begin{aligned}
D_{\text{eyes}}  &= \{d_{\text{eyes}}^{1},  d_{\text{eyes}}^{2},  \dots, d_{\text{eyes}}^{n}\},\\
D_{\text{nose}}  &= \{d_{\text{nose}}^{1},  d_{\text{nose}}^{2},  \dots, d_{\text{nose}}^{n}\},\\
&\dots \\
D_{\text{mouth}} &= \{d_{\text{mouth}}^{1}, d_{\text{mouth}}^{2}, \dots, d_{\text{mouth}}^{n}\}.
\end{aligned}
\]

Each description \( d_i \) is encoded into a textual feature vector \( v_i \) via the text encoder \(E_T(\cdot)\) of CLIP. For a given facial image \( I \), we extract its image feature \( I_{\text{feat}} \) using the image encoder \(E_I(\cdot)\) of CLIP. Then, for each region-specific set \( D_{\text{region}} \), we compute the cosine similarity between \( I_{\text{feat}} \) and every \( v_i \) in that set:
\begin{equation}
\text{sim}(I_{\text{feat}}, v_i) = \frac{I_{\text{feat}} \cdot v_i}{\|I_{\text{feat}}\| \|v_i\|}
\end{equation}
We select the text feature with the highest similarity as the most relevant description for that region:
\begin{equation}
k_{\text{region}} = \arg \max_i \, \text{sim}(I_{\text{feat}}, v_i)
\end{equation}

Finally, we concatenate the most relevant textual features across all regions to form the full semantic representation \( s \) of the face image, \( s \) serves as the region-wise aggregated semantic embedding for the image, capturing facial semantic attribute from different facial regions.

\noindent\textbf{Adapter.} We train a lightweight template$\to$attribute alignment (TAA) adapter: \(\mathcal{T} \rightarrow \mathcal{S}\) to approximate \(s\) from the facial template \(t\), \(TAA\) consists of two fully-connected layers with hidden size \(h\) and a non-linearity ReLU. The loss combines element-wise MSE and cosine alignment:
\[
\mathcal{L}_{\text{sem}} = \lambda_{\text{mse}} \| s - \hat{s} \|_2^2 + \lambda_{\text{cos}} \big( 1 - \cos(s, \hat{s}) \big),
\]
with \(\lambda_{\text{mse}}=0.7\), \(\lambda_{\text{cos}}=0.3\). Optimization uses Adam (lr \(=10^{-3}\)) for 20 epochs. After training, only \(TAA\) (not the image) is needed at attack time.

\subsection{Fusion Mapping to Latent}

After obtaining the facial template \(t\) and its predicted attribute embedding \(\hat{s}\), we concatenate them with a noise vector \(n \sim \mathcal{N}(0,I)\) and feed the fused representation \([n,\,t,\,\hat{s}]\) into the projector \(M_{\text{FLP}}\), which maps it to a latent code \(\hat{w}\in\mathcal{W}\). To make \(\hat{w}\) conform to the Style\-GAN prior, \(M_{\text{FLP}}\) is jointly trained with a critic under a WGAN objective; the full training pipeline for the \(M_{\text{FLP}}\) and critic pair is illustrated in Figure.~\ref{fig:fusion2latent}.

\noindent\textbf{The architecture of \(M_{\text{FLP}}\).}
We train the fusion mapping network \(M_{\text{FLP}}\) to project leaked facial template and semantic embedding plus controlled stochastic variation into the StyleGAN3 \(\mathcal{W}\) space. To effectively combine facial template and semantic embedding, we design \(M_{\text{FLP}}\) with a multi-branch feedforward architecture. Given a facial template \(t\), its predicted semantic embedding \(\hat{s}\) which obtained by TAA, we first obtain projected representations \( t'\) and \( s' \). The semantic embedding \( \hat{s} \) encodes facial feature attributes (eyes / nose / mouth / etc.); thus \( s' \) is partitioned into tokens \(\{s'_r\}_{r=1}^{R}\). Since facial template \(t\) encodes highly discriminative identity information. We apply multi-head attention (MHA) with the template projection as query:
\[
\tilde{s} = \mathrm{MHA}\big(Q = t',\; K = [s'_1,\ldots,s'_R],\; V = [s'_1,\ldots,s'_R]\big),
\]
producing an attention-refined semantic summary \(\tilde{s}\) emphasizing semantic embeddings most relevant to identity (e.g., eye shape, lip contour). The random vector \( n \) is normalized and directly used to provide stochastic variation. All three representations, the normalized noise vector, the projected template \(t'\), and the attention-refined semantic summary \(\tilde{s}\), are concatenated to form a unified representation \(z_f\), which is then passed through a feedforward stack (MLP + LeakyReLU blocks) to obtain a StyleGan latent:
\[
\hat{w} = M_{\text{FLP}}(n, t, \hat{s}) \in \mathcal{W}.
\]
Compared to previous methods mapping to \(\mathcal{W}\) space using \(t\) only, \(M_{\text{FLP}}\) enables the network to produce high-quality latent codes that reflect both semantic attributes and identity characteristics, fine-grained attribute conditioning via attention compensates for the expressivity otherwise gained from relaxing to \(\mathcal{W^+}\), narrowing the identity-detail trade-off. The architecture of the \(M_{\text{FLP}}\) is illustrated in Figure.~\ref{fig:FLP}.

\subsection{Training Details}

\noindent\textbf{Latent distribution alignment.}
To keep generated latent codes on the StyleGAN prior manifold, we use a GAN-based framework based on Wasserstein GAN (WGAN) algorithm to learn the distribution of intermediate latent space \(\mathcal{W}\) of StyleGAN model. Real latent codes are sampled as \(w_{\text{real}} = M_{\text{StyleGAN}}(z),\, z\sim P_Z\). Generated latent codes are \(\hat{w} = M_{\text{FLP}}(n,t,\hat{s})\). A critic \(C: \mathcal{W} \to \mathbb{R}\) is trained with:
\begin{align}
\mathcal{L}_C^{\text{WGAN}} &= \mathbb{E}_{w_{\text{real}}}[C(w_{\text{real}})] - \mathbb{E}_{\hat{w}}[C(\hat{w})], \\
\mathcal{L}_{M_{\text{FLP}}}^{\text{WGAN}} &= -\,\mathbb{E}_{\hat{w}}[C(\hat{w})].
\end{align}
The critic \(C\) is a 3-layer MLP with LeakyReLU activations.

\noindent\textbf{Reconstruction-guided refinement.}
Given the coarse latent \(\hat{w}\) predicted by \(M_{\text{FTI}}\), the frozen generator produces a face image \(\hat{I}=G(\hat{w})\). We refine \(\hat{I}\) by minimising a composite loss that balances pixel fidelity, identity consistency, attribute agreement, and perceptual realism.
\[
\mathcal{L}_{\text{rec}} =
\lambda_{\text{pix}}\mathcal{L}_{\text{pix}} +
\lambda_{\text{id}}\mathcal{L}_{\text{id}} +
\lambda_{\text{attr}}\mathcal{L}_{\text{attr}} +
\lambda_{\text{perc}}\mathcal{L}_{\text{lpips}},
\]
The final training objective therefore becomes
\[
\mathcal{L}_{M_{\text{FLP}}}^{\text{total}} =
\mathcal{L}_{M_{\text{FLP}}}^{\text{WGAN}} + \mathcal{L}_{\text{rec}}.
\]
which jointly preserves latent realism and tightens the match between \(\hat{I}\) and the ground-truth facial attributes.

\section{Implementation Details}
We employ the Adam optimizer\,\cite{kingma2017adammethodstochasticoptimization} with an initial learning rate of $1\times10^{-1}$ and a \texttt{StepLR} scheduler that multiplies the learning rate by $0.5$ every three epochs.
The total loss is a weighted sum with coefficients $\lambda_{\text{pix}} = 1.0$, $\lambda_{\text{id}} = 1$, $\lambda_{\text{attr}} = 1$ and $\lambda_{\text{lpips}} = 1$.
All experiments are conducted on a single NVIDIA RTX 3090 (24\,GB). For the StyleGAN model, we use a pretrained model of StyleGAN3 in our experiments and generate 1024$\,\times\,$1024 resolution face images.

\section{Experiments}

\noindent\textbf{Datasets.}
Following a black-box threat model, our face reconstruction network CLIP-FTI is trained on FlickrFaces-HQ (FFHQ)~\cite{karras2019style} dataset, which contains 70,000 high-quality unlabeled images. We randomly split FFHQ into 90\% training and 10\% test sets. For evaluation, we use three datasets: LFW~\cite{karras2019style}, CelebA-HQ~\cite{karras2017progressive}, and AgeDB~\cite{moschoglou2017agedb}. LFW contains 13,233 images of 5,749 subjects; CelebA-HQ includes 30,000 aligned celebrity faces at \(1024\times1024\) resolution; and AgeDB offers 16,488 images of 568 individuals with large age variations (up to \(\sim\!50\) years per identity).

\noindent\textbf{Models.}
In our experiments, we consider different SOTA FR models including ArcFace~\cite{deng2019arcface}, ElasticFace~\cite{boutros2022elasticface} as well as three different FR models with SOTA backbones from FaceX-Zoo~\cite{wang2021facex}, including HRNet~\cite{wang2020deep}, AttentionNet~\cite{wang2017residual}, and GhostNet~\cite{han2020ghostnet}. All FR models are trained on MS-Celeb1M~\cite{guo2016ms}.

\noindent\textbf{Metrics.}
For evaluation with each of the LFW, CelebA-HQ and AgeDB datasets, we build a separate FR system, and use the reference templates (i.e., enrolled in the system's database) as input to our face reconstruction method. Then we inject the reconstructed face image as a query to the system and evaluate four complementary criteria:

\medskip
\noindent\textbf{(i) Verification Protocol.}
We use BLUFR~\cite{liao2014benchmark} to verify the accuracy of identification. A gallery \(\mathcal{G}\) of enrolled templates is fixed. Cosine similarity between \(\ell_2\)-normalized embeddings provides the match score. The decision threshold is tuned on impostors to meet FAR = \{0.01\%, 0.1\%, 1\%\}; we then report TAR at the same points.
\begin{itemize}[nosep,leftmargin=*,labelsep=4pt]
    \item \textbf{Type-I} (\(P_I\)): reconstructions of the \emph{enrolled} images.  
    \item \textbf{Type-II} (\(P_{II}\)): reconstructions from a different image of the same identity.  
    \item \textbf{Impostor} (\(P_N\)): images whose identities are absent from \(\mathcal{G}\) (used only for threshold setting).
\end{itemize}

We extend BLUFR to CelebA-HQ and AgeDB by forming cross-identity (impostor) pairs via the Cartesian product, ensuring robust impostor statics and BLUFR-style evaluation. Then we report Type-I and Type-II TAR on each dataset and model at three FAR levels. The details are provided in the supplementary materials.

\smallskip
\noindent\textbf{(ii) MS-SSIM.} 
Multiscale Structural Similarity (MS{\text-}SSIM)~\cite{wang2003multiscale} evaluates luminance, contrast, and structural agreement across a cascade of down-sampled resolutions, yielding a holistic perceptual score that correlates well with human judgments on both low and high frequency content. Because it penalises structural distortions at every scale, MS{\text-}SSIM is particularly suited for assessing the global realism of GAN-based reconstructions. Values lie in \([0,1]\); larger is better.  

\smallskip
\noindent\textbf{(iii) LPIPS.} 
Learned Perceptual Image Patch Similarity (LPIPS)~\cite{zhang2018unreasonable} measures the $\ell_2$ distance between deep features extracted from a fixed, pretrained network, producing a scalar in \([0,1]\) where smaller is better.

\smallskip
\noindent\textbf{(iv) FAMSE.} 
Facial Attribute Mean-Squared Error (FAMSE) quantifies fine-grained semantic fidelity. For each image pair \((x,x')\) we first localise eyes, nose, and mouth via 68-point landmarks~\cite{zhang2016joint}; a lightweight attribute encoder then predicts a 512-D semantic vector for every region. The MSE between corresponding vectors is averaged over the three regions to obtain FAMSE, with lower scores signalling closer alignment in component-level attributes and, by extension, stronger identity consistency.

\begin{table*}[t]
\small
\setlength{\tabcolsep}{1.5pt}
\renewcommand{\arraystretch}{1.15}
\centering
\begin{tabular}{llcccccccc}
\toprule[2pt]
\textbf{Method} & $F_{\text{database}}\,/\,F_{\text{loss}}$ &
\multicolumn{4}{c}{\textbf{LFW}} &
\multicolumn{2}{c}{\textbf{CelebA-HQ}} &
\multicolumn{2}{c}{\textbf{AgeDB}} \\
\cmidrule(lr){3-6}\cmidrule(lr){7-8}\cmidrule(lr){9-10}
& &
\multicolumn{2}{c}{TAR@FAR=0.1\%} &
\multicolumn{2}{c}{TAR@FAR=1\%} &
\multicolumn{1}{c}{TAR@FAR=0.1\%} & \multicolumn{1}{c}{TAR@FAR=1\%} &
\multicolumn{1}{c}{TAR@FAR=0.1\%} & \multicolumn{1}{c}{TAR@FAR=1\%} \\
& & Type-I$\uparrow$ & Type-II$\uparrow$ & Type-I$\uparrow$ & Type-II$\uparrow$
  & Type-I$\uparrow$ & Type-I$\uparrow$ & Type-I$\uparrow$ & Type-I$\uparrow$ \\
\midrule
Otroshi \emph{et al.}
 & \multirow{2}{*}{\makecell[c]{ArcFace/\\ElasticFace}}
     & 0.9501 & 0.4655 & 0.9933 & 0.8125
     & 0.8979 & 0.9827
     & 0.7982 & 0.9405 \\
\addlinespace[-0.2em]
Ours &
     & \textbf{0.9937} & \textbf{0.8174} & \textbf{0.9995} & \textbf{0.9548}
     & \textbf{0.9535} & \textbf{0.9937}
     & \textbf{0.9002} & \textbf{0.9787} \\
Otroshi \emph{et al.}
 & \multirow{2}{*}{\makecell[c]{ElasticFace/\\ArcFace}}
     & \textbf{0.9486} & 0.5861 & \textbf{0.9893} & 0.8336
     & 0.9691 & 0.9960
     & 0.9226 & 0.9743 \\
\addlinespace[-0.2em]
Ours &
     & 0.9435 & \textbf{0.5956} & 0.9889 & \textbf{0.8517}
     & \textbf{0.9715} & \textbf{0.9967}
     & \textbf{0.9503} & \textbf{0.9898} \\
\bottomrule[1pt]
\end{tabular}
\caption{Type-I/II TAR (\%) on LFW and Type-I TAR on CelebA-HQ and AgeDB at two FAR operating points (0.1\% and 1\%). $F_{\text{database}}$ denotes the model that generated the leaked template, and $F_{\text{loss}}$ is the surrogate used during training. For CelebA-HQ and AgeDB, we only report Type-I results.}
\label{tab:mead}
\end{table*}

\begin{table*}[!t]
  \centering
  \small
  \setlength{\tabcolsep}{2.5pt}
  \begin{tabular}{l|cccc|cccc|cccc}
    \toprule[2pt]
    \multirow{2}{*}{Method} & 
    \multicolumn{4}{c|}{\textbf{LFW}} & 
    \multicolumn{4}{c|}{\textbf{CelebA-HQ}} & 
    \multicolumn{4}{c}{\textbf{AgeDB}} \\
    \cline{2-13}
     & \rule{0pt}{1.2em}MS-SSIM$\uparrow$ & LPIPS$\downarrow$ & MSE$\downarrow$ & FAMSE$\downarrow$ 
     & MS-SSIM$\uparrow$ & LPIPS$\downarrow$ & MSE$\downarrow$ & FAMSE$\downarrow$
     & MS-SSIM$\uparrow$ & LPIPS$\downarrow$ & MSE$\downarrow$ & FAMSE$\downarrow$ \\
    \midrule
    Otroshi \emph{et al.} & 0.2428 & 0.5534 & 0.0766 & 0.0503 
                          & 0.1927 & 0.5762 & 0.1014 & 0.1001 
                          & 0.2081 & 0.6064 & 0.1019 & 0.0473 \\
    Ours                  & \textbf{0.2527} & \textbf{0.5419} & \textbf{0.0662} & \textbf{0.0451} 
                          & \textbf{0.2148} & \textbf{0.5571} & \textbf{0.0880} & \textbf{0.0975} 
                          & \textbf{0.2377} & \textbf{0.5870} & \textbf{0.0884} & \textbf{0.0437} \\
    \bottomrule[1pt]
  \end{tabular}
  \caption{Quantitative comparison across datasets. $\uparrow$ indicates higher is better, $\downarrow$ lower is better.}
  \label{tab:multi_dataset_comparison}
\end{table*}

\subsection{Experimental Results}
\medskip\noindent
\textbf{Identification accuracy and attribute similarity.}
Table \ref{tab:mead} demonstrates that \emph{CLIP-FTI} consistently outperforms the strongest published inversion baseline on all three benchmarks and at both FAR operating points. In the canonical ArcFace$\rightarrow$ElasticFace setting, our method markedly elevates not only the conventional Type-I verification rate but also the more demanding Type-II metric, and these gains remain stable after the database and surrogate networks are swapped. The fact that the same pattern recurs on CelebA-HQ and AgeDB indicates that the CLIP-conditioned latent search is model-agnostic and does not overfit to any single data distribution. Further results details are included in the supplementary material.

\begin{figure}[t]
  \centering
  \includegraphics[width=\linewidth]{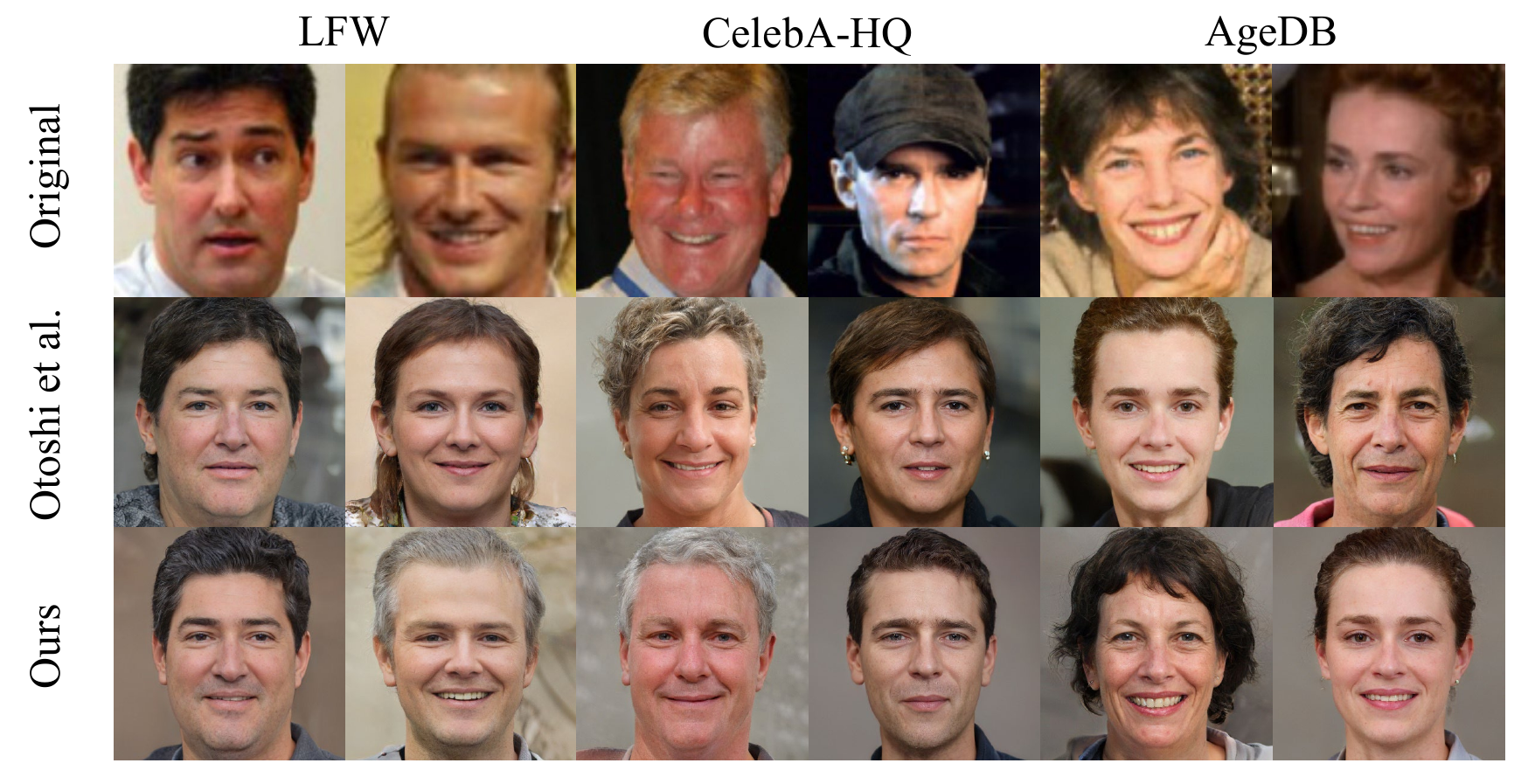}

  \caption{Qualitative comparison across datasets and reconstruction methods. The first row shows the original images, the second row shows reconstructions by \textit{Otroshi et al.}, and the third row presents the results of our CLIP-FTI.  Column groups (left to right) correspond to two examples each from LFW, CelebA-HQ, and AgeDB.}
  \label{fig:recon_grid}
\end{figure}

The qualitative evidence in Figure \ref{fig:recon_grid} echoes these quantitative results. Whereas the baseline occasionally drifts in high-level semantics (e.g., gender or overall facial attributes) \emph{CLIP-FTI} produces reconstructions that retain both global identity and fine-grained attributes across all datasets. Together, the metrics and visuals confirm that our approach obtains state-of-the-art identification accuracy while faithfully preserving the semantic attributes of reconstructed faces.

\medskip\noindent
\textbf{Component-level attribute semantics.}
As summarized in Table \ref{tab:multi_dataset_comparison}, \emph{CLIP-FTI} delivers consistent improvements over the previous best attack across all three benchmarks and every metric considered. Most notably, it reduces Facial Attribute MSE (FAMSE) (our most discriminative measure of eyes, nose, and mouth fidelity) on each dataset, indicating more accurate recovery of component-level semantics. These gains come hand-in-hand with higher MS-SSIM and lower perceptual distance scores, confirming that the method preserves or enhances global structure and visual realism while sharpening semantic details. Complementary reductions in pixel domain error further underscore the holistic quality of the reconstructions. The consistently positive margins observed across all benchmarks underscore the robustness of the CLIP-guided latent search.
Collectively, the results demonstrate that injecting textual attribute embeddings into the inversion pipeline yields faces that are both perceptually faithful and semantically precise, thereby strengthening the practical effectiveness of template-only attacks and amplifying the attendant privacy risks.

\begin{table*}[!t]
\centering
\scriptsize
\setlength{\tabcolsep}{4pt}
\renewcommand{\arraystretch}{0.92}
\setlength{\aboverulesep}{0.25ex}
\setlength{\belowrulesep}{0.25ex}
\begin{tabular}{
  C{1.9cm} C{1.9cm} C{1.9cm}
  | C{1.35cm} C{1.35cm}
  | C{1.35cm} C{1.35cm}
  | C{1.35cm} C{1.35cm}
}
\toprule
\multirow{2}{*}{\textbf{$F_{\text{database}}$}} &
\multirow{2}{*}{\textbf{$F_{\text{loss}}$}} &
\multirow{2}{*}{\textbf{$F_{\text{target}}$}} &
\multicolumn{2}{c|}{\textbf{LFW}} &
\multicolumn{2}{c|}{\textbf{CelebA-HQ}} &
\multicolumn{2}{c}{\textbf{Age-DB}} \\
& & &
{\scriptsize \textbf{Otroshi \emph{et al.}}} & \textbf{Ours} &
{\scriptsize \textbf{Otroshi \emph{et al.}}} & \textbf{Ours} &
{\scriptsize \textbf{Otroshi \emph{et al.}}} & \textbf{Ours} \\
\midrule
\multirow{5}{*}{ArcFace} & \multirow{5}{*}{ElasticFace} & ArcFace
  & 95.10 & \textbf{99.37} & 89.79 & \textbf{95.35} & 79.82 & \textbf{94.05} \\
& & ElasticFace
  & 83.90 & \textbf{91.62} & 85.31 & \textbf{91.84} & 58.04 & \textbf{68.37} \\
& & HRNet
  & 51.63 & \textbf{65.23} & 44.29 & \textbf{59.15} & 6.00 & \textbf{9.80} \\
& & AttentionNet
  & 33.99 & \textbf{43.56} & 14.26 & \textbf{23.06} & 0.96 & \textbf{2.04} \\
& & GhostNet
  & 18.72 & \textbf{27.58} & 15.22 & \textbf{22.97} & 0.97 & \textbf{1.30} \\
\midrule
\multirow{5}{*}{ElasticFace} & \multirow{5}{*}{ArcFace} & ArcFace
  & \textbf{96.41} & 96.20 & 95.22 & \textbf{95.43} & 67.16 & \textbf{69.25} \\
& & ElasticFace
  & \textbf{94.86} & 94.35 & 96.91 & \textbf{97.15} & 99.60 & \textbf{99.67} \\
& & HRNet
  & 77.35 & \textbf{77.88} & 62.99 & \textbf{64.81} & 15.66 & \textbf{18.57} \\
& & AttentionNet
  & 52.18 & \textbf{52.54} & 32.02 & \textbf{32.67} & 3.53 & \textbf{4.37} \\
& & GhostNet
  & 38.58 & \textbf{38.92} & 24.00 & \textbf{28.87} & 3.03 & \textbf{3.72} \\
\bottomrule
\end{tabular}
\caption{Cross-model transferability. Type-I TAR (\%) at FAR = $10^{-3}$ on three benchmarks for every combination of leaked backbone ($F_{\text{database}}$), surrogate backbone ($F_{\text{loss}}$), and target backbone ($F_{\text{target}}$). We compare the strongest prior attack (Otroshi \emph{et al.}) with our CLIP-FTI.}
\label{tab:bigtransfer}
\end{table*}

\begin{table}[t]
  \centering
  \scriptsize
  \setlength{\tabcolsep}{2pt}
  \renewcommand{\arraystretch}{1.05}
  \begin{tabular*}{\linewidth}{@{\extracolsep{\fill}}lcccccc@{}}
    \toprule
    Variant & Type-I$\uparrow$ & Type-II$\uparrow$ & MS-SSIM$\uparrow$ &
    MSE$\downarrow$ & FAMSE$\downarrow$ & LPIPS$\downarrow$ \\
    \midrule
    CLIP-FTI      & \textbf{0.9937} & \textbf{0.8174} & \textbf{0.2527} & \textbf{0.0662} & \textbf{0.0451} & \textbf{0.5419} \\
    w/o AttrEmb   & 0.9510 & 0.4655 & 0.2428 & 0.0766 & 0.0503 & 0.5534 \\
    w/o MHA       & 0.9553 & 0.4712 & 0.2443 & 0.0756 & 0.0501 & 0.5520 \\
    w/o ConMHA   & 0.9619 & 0.4792 & 0.2483 & 0.0742 & 0.0498 & 0.5501 \\
    \bottomrule
  \end{tabular*}
  \caption{Ablation on architectural components (\textbf{LFW}, FAR=$10^{-3}$).}
  \label{tab:ablation_module}
\end{table}
\vspace{0.6em}

\begin{table}[t]
  \centering
  \scriptsize
  \setlength{\tabcolsep}{2pt}
  \renewcommand{\arraystretch}{1.05}
  \begin{tabular*}{\linewidth}{@{\extracolsep{\fill}}lcccccc@{}}
    \toprule
    Variant & Type-I$\uparrow$ & Type-II$\uparrow$ & MS-SSIM$\uparrow$ &
    MSE$\downarrow$ & FAMSE$\downarrow$ & LPIPS$\downarrow$ \\
    \midrule
    CLIP-FTI                     & \textbf{0.9822} & \textbf{0.7229} & 0.2527 & 0.0662 & 0.0451 & 0.5419 \\
    w/o $\mathcal{L}_{\text{attr}}$ & 0.9819 & 0.6447 & \textbf{0.2563} & \textbf{0.0654} & \textbf{0.0428} & \textbf{0.5393} \\
    w/o $\mathcal{L}_{\text{lpips}}$ & 0.9618 & 0.4453 & 0.2248 & 0.0753 & 0.0450 & 0.5593 \\
    w/o both                      & 0.9783 & 0.5085 & 0.2261 & 0.0829 & 0.0495 & 0.5478 \\
    \bottomrule
  \end{tabular*}
  \caption{Ablation on loss terms (\textbf{LFW}, FAR=$10^{-3}$).}
  \label{tab:ablation_loss}
\end{table}


\medskip\noindent
\textbf{Cross-model attack transferability.}
Table~\ref{tab:bigtransfer} summarizes 30 transfer scenarios obtained by crossing two leak surrogate pairs (ArcFace$\rightarrow$ElasticFace and ElasticFace$\rightarrow$ArcFace) with five target FR and three datasets. 
\emph{CLIP-FTI} outperforms the strongest prior attack in all but two cases; in those exceptions (the ArcFace and ElasticFace self-tests under the ElasticFace$\rightarrow$ArcFace regime) our method performs virtually the same as the baseline, remaining within \(0.6\) percentage points. 
On average, it delivers double digit relative gains on the most challenging cross-architecture pairs while never regressing on any benchmark. 
Crucially, the largest gaps appear on the structurally dissimilar lightweight backbones HRNet, AttentionNet, and GhostNet, underscoring that the CLIP-conditioned latent search does not hinge on architectural similarity between the surrogate and the target. 
In summary, a single \emph{CLIP-FTI} model can successfully attack face recognition systems that differ widely in scale, training loss, and architecture, broadening the real world threat and underscoring the need for backbone-agnostic defense strategies.

\subsection{Ablation Study}

\medskip\noindent
\textbf{Effect of CLIP-driven attribute embeddings and cross-modal attention.}
Table~\ref{tab:ablation_module} yields three observations.
(i) \textit{Embeddings matter.} Removing CLIP-driven attribute embeddings (w/o AttrEmb) causes the largest drop across all metrics, showing that a template-only signal lacks the component-level semantics needed for cross-view verification.
(ii) \textit{Attention matters.} Simple concatenation (w/o MHA) recovers little, indicating the need for an explicit mechanism to adaptively weight attribute cues.
(iii) \textit{Query choice matters most.} Using MHA without the template as query (w/o ConMHA) improves over concatenation but trails the full model; making the template guide the attention is crucial for focusing on identity-determining parts (eyes, nose, mouth) and fully exploiting the attribute embeddings.

\medskip\noindent
\textbf{Effect of attribute and perceptual losses.}
Table~\ref{tab:ablation_loss} indicates that the two loss terms address complementary aspects of reconstruction.
Without either loss, the generator converges to over-smoothed images and performs poorly on all metrics.
Adding only the \emph{attribute loss} (w/o $\mathcal{L}_{\text{lpips}}$) emphasizes global traits from an identity-agnostic extractor (e.g., age, expression), lowering pixel-space errors (MSE, FAMSE) but yielding little change in Type-I/II TAR.
Applying only the \emph{perceptual loss} (w/o $\mathcal{L}_{\text{attr}}$) improves local texture/structure and verification accuracy, while fine attributes drift and pixel metrics remain largely unchanged.
Optimizing both losses balances these effects: the attribute loss provides semantic guidance, and the perceptual loss restores high-frequency detail, yielding the best verification accuracy with a minor increase in pixel error—an acceptable trade-off when the goal is to pass FR checks.

\medskip\noindent
\textbf{Effect of CLIP attribute guidance on local regions.}
To verify that our CLIP attribute embeddings act on the intended facial parts, we compute region-wise CLIP cosine similarity between (i) images produced by Otroshi et al. and (ii) images produced by ours, and the corresponding attribute embeddings (eyes, nose, mouth, jaw, eyebrow). The detail results are provided in the supplementary materials.

\section{Conclusion}
In this paper, we revisited face template inversion (FTI) through fine-grained attribute recovery and proposed \textbf{CLIP-FTI}. By predicting component-level facial attribute embeddings from a leaked template and fusing them with the template via a cross-modal attention network, and project to StyleGAN \(\mathcal{W}\) space to synthesize identity-consistent, attribute-faithful faces. Extensive experiments on multiple datasets and FR backbones show that \textbf{CLIP-FTI} achieves a new state-of-the-art across all attack metrics on standard benchmarks while retaining strong cross-model transferability. To the best of our knowledge, we are the first to introduce external textual semantic information into the face template inversion attack task, enabling more precise and controllable attribute reconstruction across diverse facial regions.

\section{Acknowledgments}
This work is supported in part by the National Natural Science Foundation of China under grant numbers, U23B2023 and 62472199, Guangdong Key Laboratory of Data Security and Privacy Preserving under Grant 2023B1212060036, the basic and Applied Basic Research Foundation of Guangdong Province (2025A1515011097), and the Outstanding Youth Project of Guangdong Basic and Applied Basic Research Foundation (2023B1515020064). This work is also supported by Engineering Research Center of Trustworthy AI, Ministry of Education.

\bibliography{main}

\clearpage

\section{Supplementary Materials}

\begin{table*}[!t]
\centering
\scriptsize 
\setlength{\tabcolsep}{3pt}        
\renewcommand{\arraystretch}{0.9}  
\setlength{\aboverulesep}{0.2ex}   
\setlength{\belowrulesep}{0.2ex}   
\caption{Results at different FAR thresholds (Type-I TAR). Left: CelebA-HQ; Right: AgeDB.}
\label{tab:celebahq_agedb_joint}
\begin{tabular*}{\textwidth}{@{\extracolsep{\fill}} l ccc ccc @{}}
\toprule
\multirow{2}{*}{Method} &
\multicolumn{3}{c}{\textbf{CelebA-HQ}} &
\multicolumn{3}{c}{\textbf{AgeDB}} \\
\cmidrule(lr){2-4}\cmidrule(lr){5-7}
& TAR@FAR=1\% & TAR@FAR=0.1\% & TAR@FAR=0.01\% & TAR@FAR=1\% & TAR@FAR=0.1\% & TAR@FAR=0.01\% \\
\midrule
ArcFace
& 0.19997557997703552 & 0.2802311182022095 & 0.35252469778060913
& 0.2073499858379364  & 0.2887561619281769 & 0.36410874128341675 \\
ElasticFace
& 0.23934301733970642 & 0.3263251483440399 & 0.39957278966903687
& 0.2750087380409241  & 0.37256500124931335 & 0.46182289719581604 \\
\bottomrule
\end{tabular*}
\end{table*}

\subsection{TAA adapter}
\noindent\textbf{Facial feature attribute prompt matching.}
To enable more detailed semantic alignment, we divide the face into multiple regions—such as eyes, nose, and mouth—and predefine a set of textual descriptions for each region:
\[
\begin{aligned}
D_{\text{eyes}}  &= \{d_{\text{eyes}}^{1},  d_{\text{eyes}}^{2},  \dots, d_{\text{eyes}}^{n}\},\\
D_{\text{nose}}  &= \{d_{\text{nose}}^{1},  d_{\text{nose}}^{2},  \dots, d_{\text{nose}}^{n}\},\\
&\dots \\
D_{\text{mouth}} &= \{d_{\text{mouth}}^{1}, d_{\text{mouth}}^{2}, \dots, d_{\text{mouth}}^{n}\}.
\end{aligned}
\]

Each description \( d_i \) is encoded into a textual feature vector \( v_i \) via the text encoder \(E_T(\cdot)\) of CLIP. For a given facial image \( I \), we extract its image feature \( I_{\text{feat}} \) using the image encoder \(E_I(\cdot)\) of CLIP. Then, for each region-specific set \( D_{\text{region}} \), we compute the cosine similarity between \( I_{\text{feat}} \) and every \( v_i \) in that set:
\begin{equation}
\text{sim}(I_{\text{feat}}, v_i) = \frac{I_{\text{feat}} \cdot v_i}{\|I_{\text{feat}}\| \|v_i\|}
\end{equation}
We select the text feature with the highest similarity as the most relevant description for that region:
\begin{equation}
k_{\text{region}} = \arg \max_i \, \text{sim}(I_{\text{feat}}, v_i)
\end{equation}

\begin{figure*}[t]  
  \centering
  \includegraphics[width=0.8\linewidth]{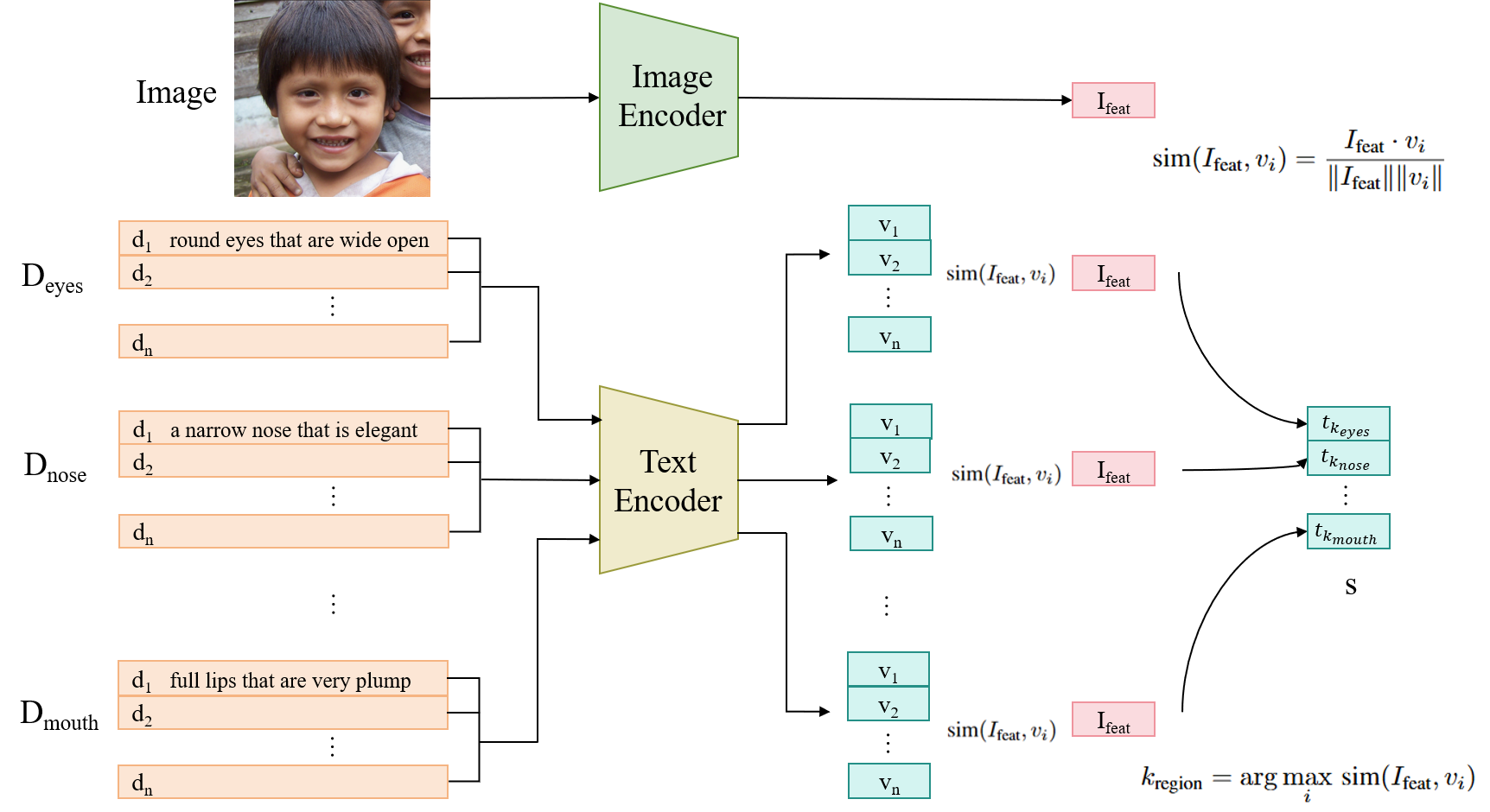}

  \caption{The process of obtaining textual semantic features involves extracting textual features for each predefined set of text content using a text encoder. For each image, image features are extracted using an image encoder. Subsequently, the feature with the minimum cosine similarity to the image features is identified from each group of textual features. These selected textual features are then concatenated to form the textual semantic feature for the corresponding image.}
  \label{fig:embeddings}
\end{figure*}

Finally, we concatenate the most relevant textual features across all regions to form the full semantic representation \( s \) of the face image, \( s \) serves as the region-wise aggregated semantic embedding for the image, capturing facial semantic attribute from different facial regions. Figure~\ref{fig:embeddings} shows the process of obtaining textual semantic features.

\subsection{Experiments}

\noindent\textbf{Metrics.}

\medskip

\noindent\textbf{(i) Verification Protocol.}
We use BLUFR~\cite{liao2014benchmark} to verify the accuracy of identification. A gallery \(\mathcal{G}\) of enrolled templates is fixed. Cosine similarity between \(\ell_2\)-normalized embeddings provides the match score. The decision threshold is tuned on impostors to meet FAR = \{0.01\%, 0.1\%, 1\%\}; we then report TAR at the same points.
\begin{itemize}[nosep,leftmargin=*,labelsep=4pt]
    \item \textbf{Type-I} (\(P_I\)): reconstructions of the \emph{enrolled} images.  
    \item \textbf{Type-II} (\(P_{II}\)): reconstructions from a different image of the same identity.  
    \item \textbf{Impostor} (\(P_N\)): images whose identities are absent from \(\mathcal{G}\) (used only for threshold setting).
\end{itemize}

We extend BLUFR to CelebA-HQ and AgeDB by forming cross-identity (impostor) pairs via the Cartesian product, ensuring robust impostor statics and BLUFR-style evaluation. Then we report Type-I and Type-II TAR on each dataset and model at three FAR levels. Table \ref{tab:celebahq_agedb_joint} reports the operating thresholds corresponding to TAR@FAR = {1\%, 0.1\%, 0.01\%} on CelebA-HQ and AgeDB under the ArcFace and ElasticFace models.”

\subsection{Experimental Results}

\medskip\noindent
\textbf{Identification accuracy and attribute similarity.}
Table~\ref{tab:mead_simplified} in the supplementary material further substantiates the superior cross-model generalization capacity of \emph{CLIP-FTI}. Compared with the strongest published baseline, our method consistently achieves higher Type-I and Type-II verification performance across diverse face recognition backbones and datasets. Notably, the performance advantages of \emph{CLIP-FTI} persist even under challenging cross-architecture configurations where the database and surrogate models are intentionally mismatched. This robustness, observed on both LFW and out-of-distribution benchmarks such as CelebA-HQ and AgeDB, highlights the model-agnostic nature of our CLIP-guided latent space traversal and reinforces its resilience against overfitting to a specific identity embedding scheme.

\begin{figure}[t]  
  \centering
  \includegraphics[width=\linewidth]{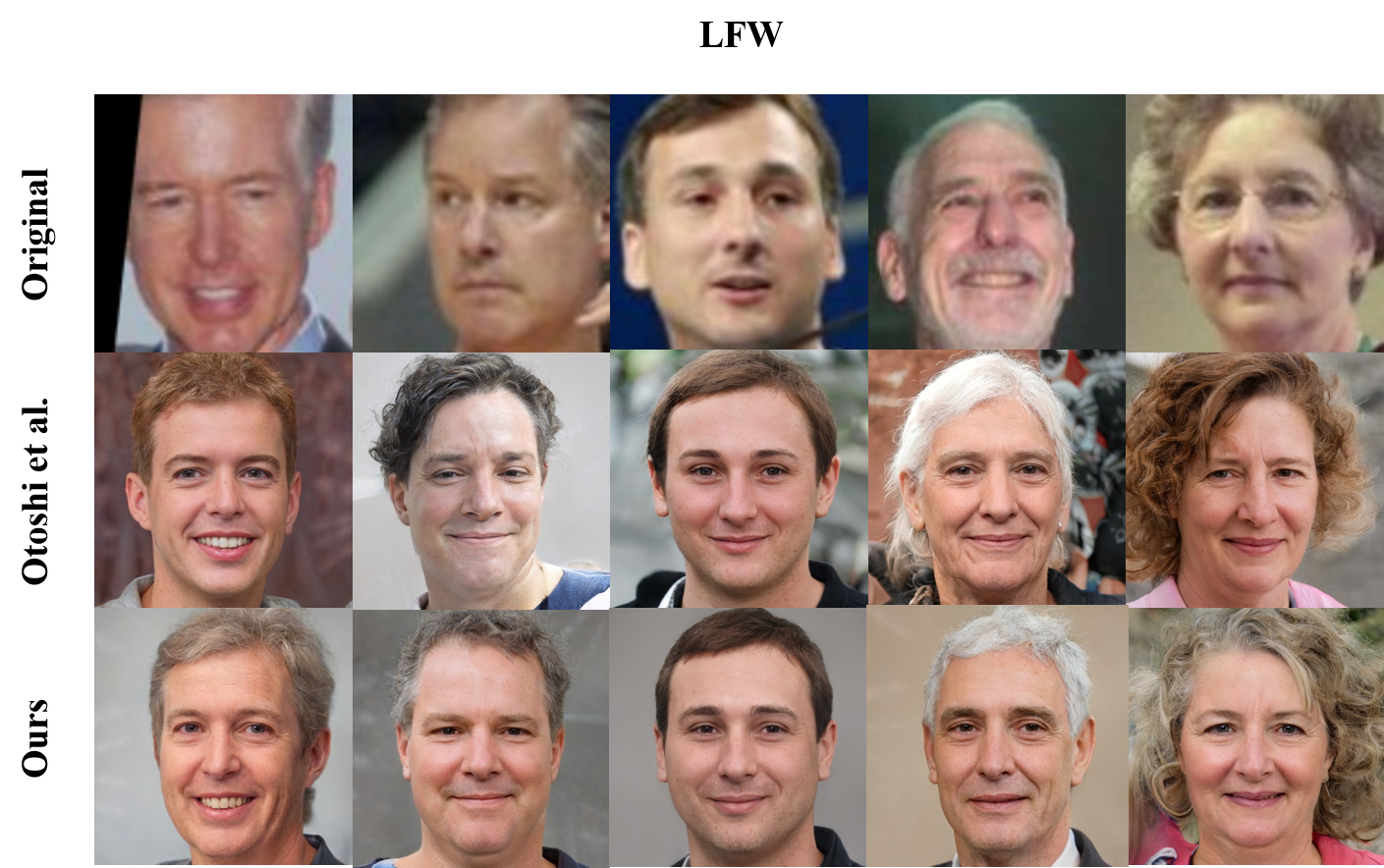}

  \caption{Qualitative comparison on LFW datasets and reconstruction methods. The first row shows the original images, the second row shows reconstructions by \textit{Otroshi et al.}, and the third row presents the results of our CLIP-FTI.}
  \label{fig:LFW}
\end{figure}

\begin{figure}[t]  
  \centering
  \includegraphics[width=\linewidth]{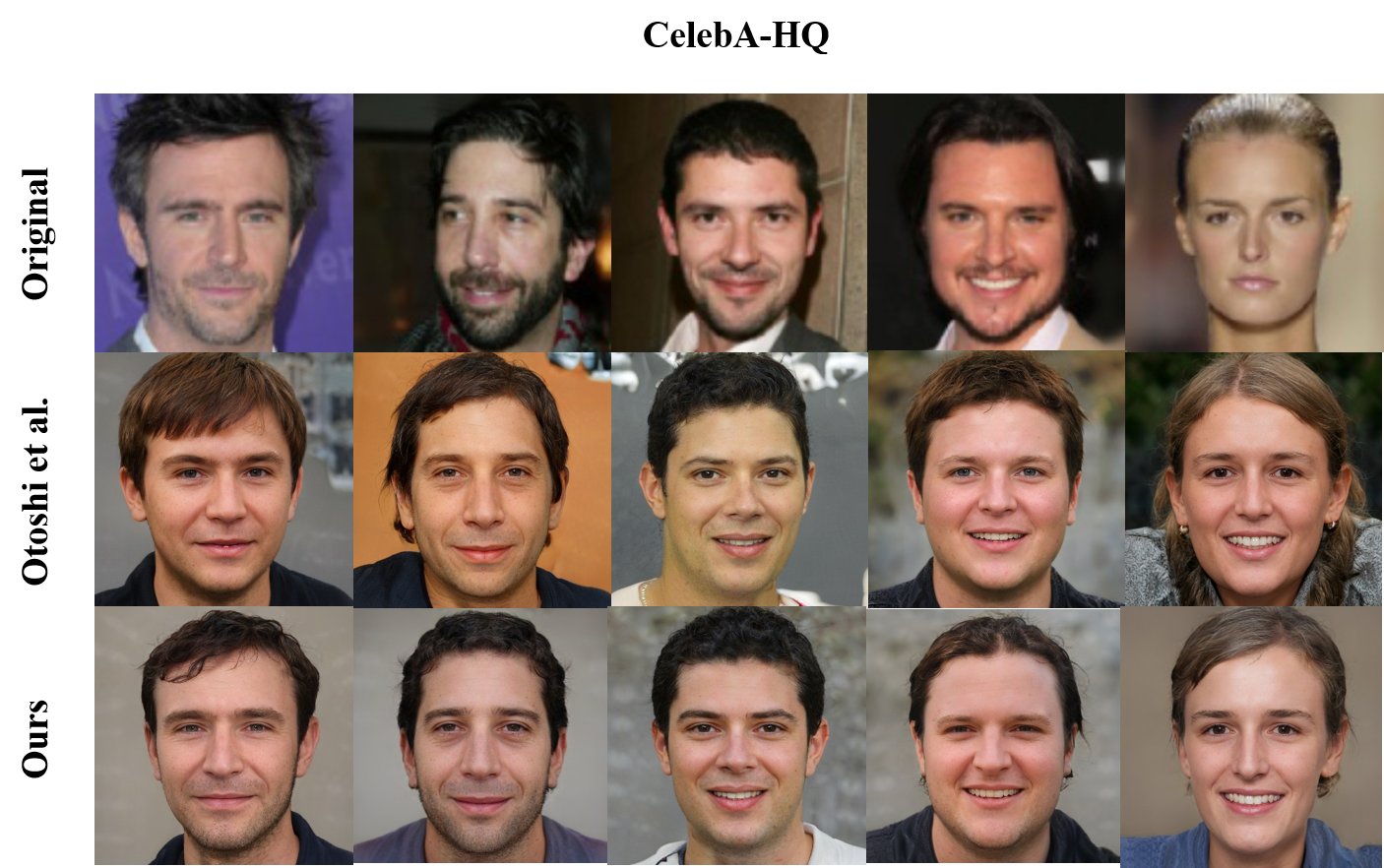}

  \caption{Qualitative comparison on CelebA-HQ datasets and reconstruction methods. The first row shows the original images, the second row shows reconstructions by \textit{Otroshi et al.}, and the third row presents the results of our CLIP-FTI.}
  \label{fig:CelebA-HQ}
\end{figure}

\begin{figure}[t]  
  \centering
  \includegraphics[width=\linewidth]{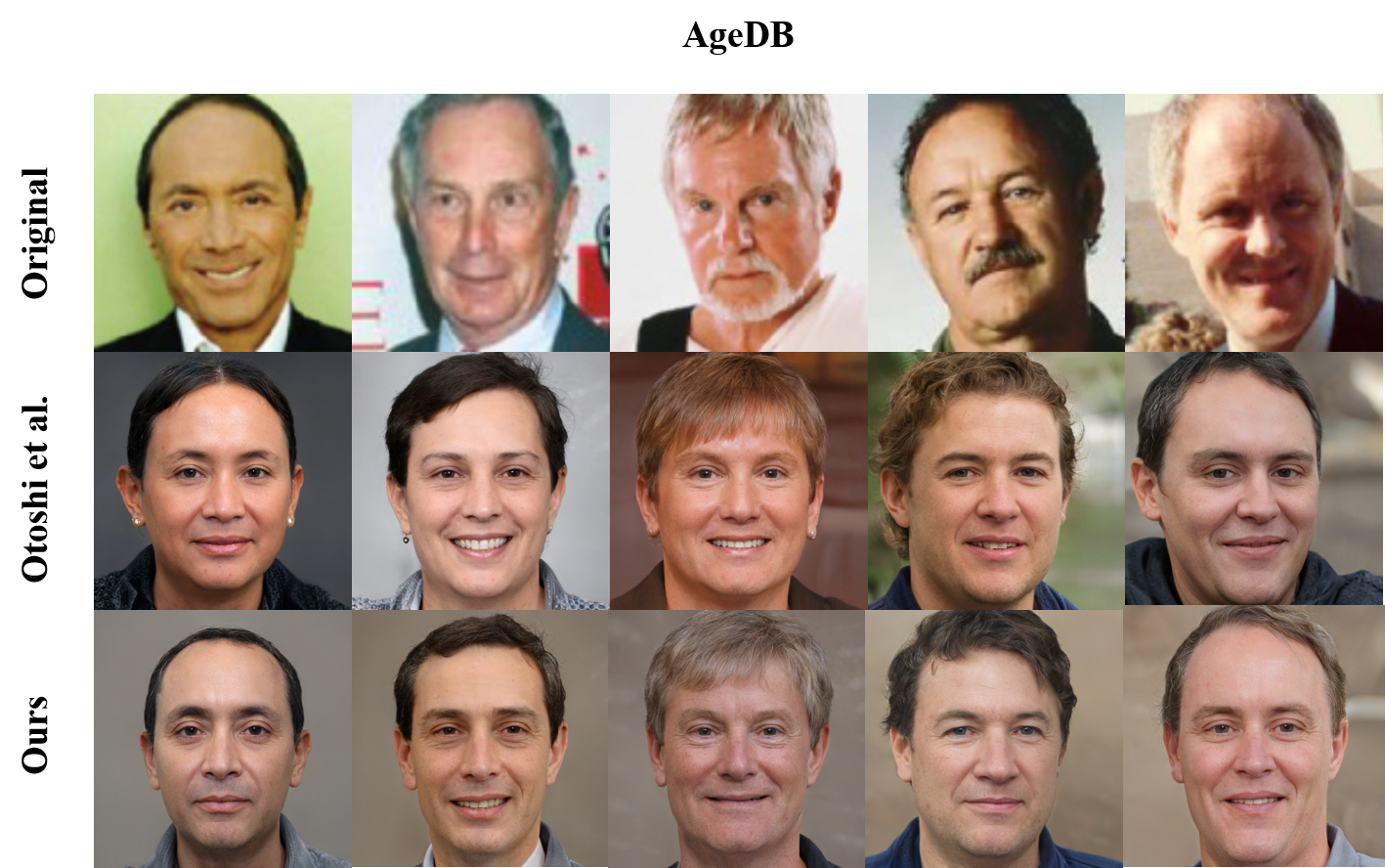}

  \caption{Qualitative comparison on AgeDB datasets and reconstruction methods. The first row shows the original images, the second row shows reconstructions by \textit{Otroshi et al.}, and the third row presents the results of our CLIP-FTI.}
  \label{fig:AgeDB}
\end{figure}

The qualitative results presented in Figure~\ref{fig:LFW},~\ref{fig:CelebA-HQ} and~\ref{fig:AgeDB}, (now organized into separate visualizations per dataset) reinforce the quantitative trends. For each dataset, we display representative reconstructions across five identities, with columns corresponding to different inversion methods. Compared to the baseline, which occasionally exhibits semantic drift—such as altered gender cues or inconsistent facial structure. \emph{CLIP-FTI} consistently preserves both the global appearance and subtle identity attributes. Across LFW, CelebA-HQ, and AgeDB, the reconstructions produced by our method exhibit sharper semantic alignment with the target features, particularly in regions critical for recognition such as the eyes, mouth, and facial contour. These consistent visual gains, when viewed alongside the numerical results, further validate the superiority of our CLIP-conditioned inversion pipeline in achieving perceptually and semantically faithful reconstructions.

\begin{table}[h]
\small
\setlength{\tabcolsep}{2pt}
\renewcommand{\arraystretch}{1.15}
\centering
\caption{Type-I/II TAR (\%) on LFW, and Type-I TAR on CelebA-HQ and AgeDB. $F_{\text{database}}$ denotes the model that generated the leaked template, and $F_{\text{loss}}$ is the surrogate used during training. All results are reported at FAR=0.01\%.}
\label{tab:mead_simplified}
\begin{tabular}{llccccc}
\toprule[2pt]
\multirow{2}{*}{\makecell[c]{\textbf{Method}}} &  
\multirow{2}{*}{\makecell[c]{$F_{\text{database}}$/\\$F_{\text{loss}}$}}&
\multicolumn{2}{c}{\makecell{\textbf{LFW}}} &
\multicolumn{1}{c}{\makecell{\textbf{CelebA-HQ}}} &
\multicolumn{1}{c}{\makecell{\textbf{AgeDB}}} \\
\cmidrule(lr){3-4} \cmidrule(lr){5-5} \cmidrule(lr){6-6}
& & Type-I$\uparrow$ & Type-II$\uparrow$ & Type-I$\uparrow$ & Type-I$\uparrow$ \\
\midrule
Otroshi \emph{et al.} &
\multirow{2}{*}{\makecell[c]{ArcFace/\\ElasticFace}} &
0.8326 & 0.1779 & 0.6916 & 0.5376 \\
Ours &
& \textbf{0.9556} & \textbf{0.5171} & \textbf{0.8162} & \textbf{0.6919} \\
\addlinespace
Otroshi \emph{et al.} &
\multirow{2}{*}{\makecell[c]{ElasticFace/\\ArcFace}} &
\textbf{0.8456} & 0.3421 & 0.8165 & 0.8165 \\
Ours &
& 0.8268 & \textbf{0.3540} & \textbf{0.8909} & \textbf{0.8550} \\
\bottomrule[1pt]
\end{tabular}
\end{table}

\medskip\noindent
\textbf{Component-level attribute semantics.}
Table~\ref{tab:multi_dataset_comparison2} presents a comprehensive evaluation of the reconstructed image quality under a cross-model configuration, where the template generator and inversion surrogate are drawn from different face recognition backbones (ElasticFace and ArcFace). Under this challenging setting, \emph{CLIP-FTI} consistently surpasses the strongest prior baseline across all metrics and datasets. The improvements span structural similarity, perceptual alignment, and pixel-level reconstruction fidelity, highlighting the model's ability to preserve both global facial structure and fine-grained visual realism.

Crucially, the method achieves lower facial attribute reconstruction errors, reflecting more accurate alignment of identity-relevant components such as the eyes, nose, and mouth. These improvements in FAMSE (our most attribute-sensitive measure) suggest that the CLIP-guided latent optimization not only refines visual detail but also preserves identity semantics in regions most critical to recognition. Taken together, these results affirm that injecting textual priors into the inversion process strengthens the semantic consistency of reconstructions even under architectural mismatch, further expanding the practical reach and privacy implications of template inversion attacks.

\begin{table*}[!t]
  \centering
  \small
  \setlength{\tabcolsep}{3pt}
  \caption{Quantitative comparison across datasets under (ElasticFace / ArcFace) configuration. $\uparrow$ indicates higher is better, $\downarrow$ lower is better.}
  \label{tab:multi_dataset_comparison2}
  \begin{tabular}{l|cccc|cccc|cccc}
    \toprule[2pt]
    \multirow{2}{*}{Method} & 
    \multicolumn{4}{c|}{\textbf{LFW}} & 
    \multicolumn{4}{c|}{\textbf{CelebA-HQ}} & 
    \multicolumn{4}{c}{\textbf{AgeDB}} \\
    \cline{2-13}
     & \rule{0pt}{1.2em}MS-SSIM$\uparrow$ & LPIPS$\downarrow$ & MSE$\downarrow$ & FAMSE$\downarrow$ 
     & MS-SSIM$\uparrow$ & LPIPS$\downarrow$ & MSE$\downarrow$ & FAMSE$\downarrow$
     & MS-SSIM$\uparrow$ & LPIPS$\downarrow$ & MSE$\downarrow$ & FAMSE$\downarrow$ \\
    \midrule
    Otroshi \emph{et al.} & 0.2377 & 0.5389 & 0.0833 & 0.0543 
                          & 0.2256 & 0.5735 & 0.1012 & 0.0472 
                          & 0.1989 & 0.6031 & 0.1076 & 0.1101 \\
    Ours                  & \textbf{0.2632} & \textbf{0.5350} & \textbf{0.0677} & \textbf{0.0496} 
                          & \textbf{0.2387} & \textbf{0.5563} & \textbf{0.0907} & \textbf{0.0457} 
                          & \textbf{0.2283} & \textbf{0.5836} & \textbf{0.0851} & \textbf{0.1059} \\
    \bottomrule[1pt]
  \end{tabular}
\end{table*}

\begin{table}[t]
  \centering
  \scriptsize
  \setlength{\tabcolsep}{2pt}
  \renewcommand{\arraystretch}{1.05}
  \caption{Ablation Study on Conditional Attention and Multi-Head Configuration (\textbf{LFW}, FAR=$10^{-3}$).}
  \label{tab:ablation_module2}
  \begin{tabular*}{\linewidth}{@{\extracolsep{\fill}}lcccccc@{}}
    \toprule
    Variant & Type-I$\uparrow$ & Type-II$\uparrow$ & MS-SSIM$\uparrow$ &
    MSE$\downarrow$ & FAMSE$\downarrow$ & LPIPS$\downarrow$ \\
    \midrule
    CLIP-FTI      & \textbf{0.9937} & \textbf{0.8174} & \textbf{0.2527} & \textbf{0.0662} & \textbf{0.0451} & \textbf{0.5419} \\
    MHA          & 0.9619 & 0.4792 & 0.2483 & 0.0742 & 0.0498 & 0.5501 \\
    MHA (head=4) & 0.9569 & 0.4666 & 0.2328 & 0.0707 & 0.0468 & 0.5602 \\
    ConMHA       & 0.9783 & 0.5085 & 0.2261 & 0.0829 & 0.0495 & 0.5477 \\
    \bottomrule
  \end{tabular*}
\end{table}
\vspace{0.6em}

\begin{table}[t]
  \centering
  \scriptsize
  \setlength{\tabcolsep}{2pt}
  \renewcommand{\arraystretch}{1.05}
  \caption{Ablation on loss terms (\textbf{LFW}, FAR=$10^{-3}$).}
  \label{tab:ablation_loss2}
  \begin{tabular*}{\linewidth}{@{\extracolsep{\fill}}lcccccc@{}}
    \toprule
    Variant & Type-I$\uparrow$ & Type-II$\uparrow$ & MS-SSIM$\uparrow$ &
    MSE$\downarrow$ & FAMSE$\downarrow$ & LPIPS$\downarrow$ \\
    \midrule
    MHA  & 0.9619 & 0.4792 & 0.2483 & 0.0742 & 0.0498 & 0.5501 \\
    MHA + $\mathcal{L}_{\text{attr}}$ & 0.9764 & 0.5566 & 0.2367 & 0.0736 & 0.0478 & 0.5532 \\
    MHA + 4$\mathcal{L}_{\text{attr}}$ & 0.9754 & 0.5248 & 0.2251 & 0.0786 & 0.0498 & 0.5601 \\
    MHA + $\mathcal{L}_{\text{lpips}}$ & 0.9841 & 0.6957 & 0.2604 & 0.0657 & 0.0433 & 0.5636 \\
    MHA + $\mathcal{L}_{\text{lpips}}$ + $\mathcal{L}_{\text{attr}}$ & 0.9833 & 0.7148 & 0.2622 & 0.0669 & 0.0456 & 0.5628 \\
    MHA + $\mathcal{L}_{\text{lpips}}$ + 4$\mathcal{L}_{\text{attr}}$ & 0.9822 & 0.7060 & 0.2562 & 0.0698 & 0.0454 & 0.5598 \\
    \bottomrule
  \end{tabular*}
\end{table}

\begin{table}[t]
  \centering
  \scriptsize
  \setlength{\tabcolsep}{2pt}
  \renewcommand{\arraystretch}{1.05}
  \caption{CLIP similarity on corresponding regions between Otroshi \emph{et al.} and ours.}
  \label{tab:interpretability}
  \begin{tabular*}{\linewidth}{@{\extracolsep{\fill}}lccccc@{}}
    \toprule
    Method & eyes & nose & mouth & jaw & eyebrow \\
    \midrule
    Otroshi \emph{et al.}  & 0.1510 & 0.1772 & 0.1725 & 0.1778 & 0.1697  \\
    Ours & \textbf{0.2087} & \textbf{0.2364} & \textbf{0.2305} & \textbf{0.2350} & \textbf{0.2252}  \\
    \bottomrule
  \end{tabular*}
\end{table}

\subsection{Ablation Study}

\medskip\noindent
\textbf{Effect of Conditional Attention.}
Table~\ref{tab:ablation_module2} investigates the impact of different architectural variants of our attention mechanism. Three key observations emerge:

(i) \textit{Multi-head configuration is non-trivial.} Introducing multiple attention heads without sufficient semantic diversity, such as using four heads under the current attribute design, leads to degradation in both Type-I and Type-II verification accuracy. While minor improvements can be observed in certain perceptual or pixel-level metrics, these gains come at the expense of semantic alignment, suggesting that excessive head splitting may dilute the attribute signal across insufficiently differentiated facial regions.

(ii) \textit{Single-head attention with conditional guidance is essential.} Compared to naive multi-head variants, our conditional attention design (ConMHA), which explicitly uses the template as the query to guide feature selection, substantially improves identity verification accuracy. Although this variant alone incurs slight regressions in global appearance metrics, such drops are mitigated in the full model via auxiliary loss designs, validating the necessity of query-aware attention for semantic precision.

(iii) \textit{CLIP-FTI benefits from tight attention-attribute coupling.} The full model, which combines conditional attention with CLIP-driven embeddings, achieves consistent gains across all objectives. These results indicate that accurate semantic reconstruction depends not only on the availability of attribute priors but also on architectural choices that allow the network to leverage them in an identity-sensitive manner.

\medskip\noindent
\textbf{Effect of Attribute and Perceptual Losses.}
Table~\ref{tab:ablation_loss2} analyzes the contribution of different loss components to the overall performance of the inversion model. Introducing the attribute loss yields consistent improvements in both Type-I and Type-II verification rates, confirming its utility in guiding semantic alignment. However, increasing its weight leads to a decline in recognition accuracy, likely due to the overemphasis on coarse attributes at the expense of identity-specific details. In contrast, adding the perceptual loss (\(\mathcal{L}_{\text{lpips}}\)) leads to substantial and robust gains across both recognition metrics and perceptual quality measures, indicating its effectiveness in preserving both global structure and local fidelity.

Combining the attribute and perceptual losses results in further gains in identity preservation, yet again reveals a sensitivity to the attribute loss weight: aggressive weighting introduces diminishing returns or even regressions in verification performance. To strike a practical balance between reconstruction fidelity and attack success, we adopt a default configuration where both \(\mathcal{L}_{\text{attr}}\) and \(\mathcal{L}_{\text{lpips}}\) are included with equal weights, ensuring stable and high-quality reconstructions under the attack scenario.

\medskip\noindent
\textbf{Effect of CLIP attribute guidance on local regions.}
To verify that our CLIP attribute embeddings act on the intended facial parts, we compute region-wise CLIP cosine similarity between (i) images produced by Otroshi et al. and (ii) images produced by ours, and the corresponding attribute embeddings (eyes, nose, mouth, jaw, eyebrow). Images are encoded with the CLIP image encoder; similarities are averaged over a held-out set. As shown in Table~\ref{tab:interpretability}, our method achieves uniformly higher alignment on all five parts. These consistent gains indicate that CLIP guidance concentrates on the semantically matched local regions.

\end{document}